\documentclass[10pt,twocolumn,letterpaper]{article}

\usepackage{cvpr}

\definecolor{darkgreen}{RGB}{0, 128, 0}

\usepackage{multirow}

\usepackage{tikz}
\usetikzlibrary{fit, positioning, shapes.misc} 

\usepackage{siunitx}
\newcommand{\EMParticles}{EMParticles\xspace}
\newcommand{\LiveCell}{LiveCell\xspace}
\newcommand{\OrganoidBasic}{OrganoidBasic\xspace}
\newcommand{\KvasirSEG}{KvasirSEG\xspace}

\newcommand{\samed}{SAMed\xspace}
\newcommand{\autosam}{AutoSAM\xspace}
\newcommand{\cellseg}{CellSeg1\xspace}
\newcommand{\nnunet}{nnU-Net\xspace}

\definecolor{cvprblue}{rgb}{0.21,0.49,0.74}
\usepackage[pagebackref,breaklinks,colorlinks,allcolors=cvprblue]{hyperref}

\def\paperID{*****}
\def\confName{CVPR}
\def\confYear{2025}

\title{Prompt-Tuning SAM: From Generalist to Specialist\\ with only 2,048 Parameters and 16 Training Images}

\author{Tristan Piater \quad Bj\"{o}rn Barz \quad Alexander Freytag\\
Carl Zeiss AG, Germany\\
{\tt\small tristan.piater@zeiss.com, bjoern.barz@zeiss.com, alexander.freytag@zeiss.com}
}

\begin{document}

\maketitle
\begin{abstract}

The Segment Anything Model (SAM) 
is widely used for segmenting a diverse range of objects in natural images from simple user prompts like points or bounding boxes.
However, SAM's performance decreases substantially when applied to non-natural domains like microscopic imaging.
Furthermore, due to SAM's interactive design, it requires a precise prompt for each image and object, which is unfeasible in many automated biomedical applications.
Previous solutions adapt SAM by 
training millions of parameters via fine-tuning large parts of the model or of adapter layers. 
In contrast,
we show that as little as {2,048} additional parameters are sufficient for turning SAM into a use-case specialist for a certain downstream task.
Our novel PTSAM (prompt-tuned SAM) method uses prompt-tuning, a parameter-efficient fine-tuning technique, to adapt SAM for a specific task.
We validate the performance of our approach on multiple microscopic and one medical dataset.
Our results show that prompt-tuning only SAM's mask decoder already leads to a performance on-par with state-of-the-art techniques while requiring roughly {2,000}x less trainable parameters.
For addressing domain gaps, we find that additionally prompt-tuning SAM's image encoder is beneficial, further improving segmentation accuracy by up to 18\% over state-of-the-art results.
Since PTSAM can be reliably trained with as little as {16} annotated images,
we find it particularly helpful for applications with limited training data and domain shifts.

\end{abstract}

\section{Introduction}
\label{sec:intro}

\begin{figure}
    \centering
    \includegraphics[width=\linewidth]{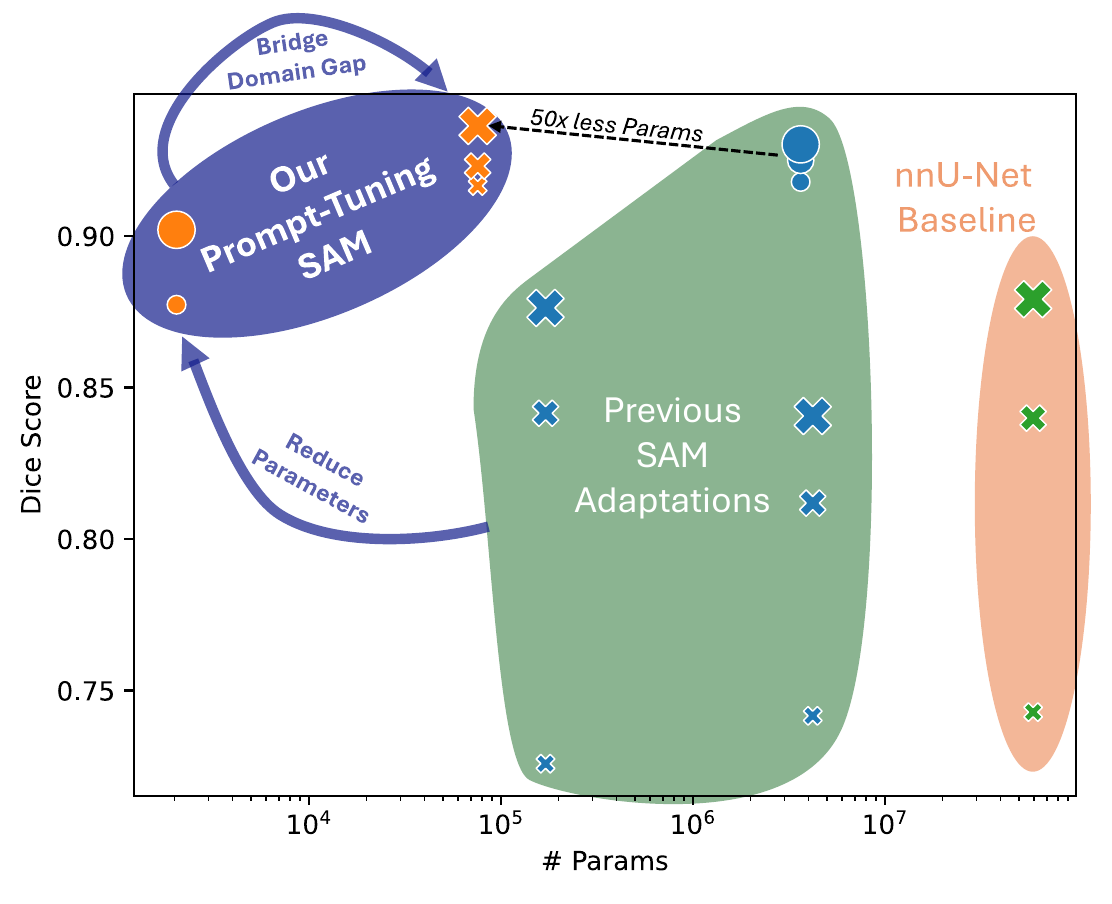}
    \caption{Comparison of Prompt-Tuning SAM (ours) to other techniques for adapting SAM, as well as  nnU-Net as the de-facto standard.
    Circles indicate adapting the mask decoder only, while crosses indicate additional adaptation of the image encoder. Marker sizes correlate with the number of training images.}
   
    \label{fig:dummy_res}
\end{figure}

Biological research relies strongly on high-throughput experiments. 
Consequently, 
high amounts of recorded data need to be reliable analyzed at scale~\cite{Akbar2019ImageAF, Shen2023CellSA,Schrter2024ALA,Anaam2023DeepAL,Singla2020ARP}.
It is therefore no surprise that automated image analysis techniques are of utmost importance.
One important technique is semantic segmentation,
which enables a broad range of solutions, 
like automatic calculation of cell mass \cite{Akbar2019ImageAF, Shen2023CellSA}, comparison of organoid volumes \cite{Schrter2024ALA}, and detection of mitotic cells \cite{Anaam2023DeepAL}.
Because applications can differ strongly, \eg,  in the required microscopic imaging modality \cite{Singla2020ARP},
or in the targets to segment, 
each application currently requires a separate model that is explicitly trained for this specific task.
However, training such a specialist model is expensive, requiring computational resources, annotated data, and time of machine learning experts.
All three aspects are usually rare resources in microscopy image analysis facilities.

In contrast to such specialized models,
the recent success of large-scale unsupervised pre-training especially for transformer architectures started a shift towards generalist models~ \cite{Vaswani2017AttentionIA,Brown2020LanguageMA,Achiam2023GPT4TR}.
These so-called foundation models 
\cite{Bommasani2021OnTO},
enabled by self-supervised pre-training without being trained for only one specific use case,
have advanced zero- and few-shot generalization \cite{Brown2020LanguageMA} to solve tasks outside of their training domain.
After having started in the language domain,  
benefits of foundation models were explored in the vision domain as well \cite{Dosovitskiy2020AnII, radford2021learning, he2021maskedautoencodersscalablevision}. 
Consequently, vision foundation models (VFM) were developed across many domains, 
\eg,
RetFound by \citet{BiccasNeto2023RETFoundTR} enabling fast training of \textit{any} retina analysis solutions,  
DepthAnything by \citet{Yang2024DepthAU} performs depth estimation for \textit{any} scene, or
UniverSeg by \citet{Butoi2023UniverSegUM}, which performs in-context segmentation of \textit{any} medical object.
In the following, we will focus on another popular VFM:  
the Segment Anything Model (SAM) by \citet{Kirillov2023SegmentA},
which enabled interactive segmentation of \textit{any} natural-scene object.

SAM can be applied to a wide range of tasks with impressive zero-shot results.
To achieve this performance, SAM was trained on SA-1B \cite{Kirillov2023SegmentA},
a huge segmentation database consisting of one billion of masks.
However, as SA-1B covers only natural images, it can not be expected a-priori that SAM generalizes beyond natural images. 
In fact, it has been shown that SAM's segmentation performance significantly drops when applied in non-natural image domains like agriculture, medicine and microscopy \cite{ji2024segment, deng2023segment, israel2023foundationmodelcellsegmentation, archit2025segment, Tsiporenko2024GoingBU}.
Furthermore, 
SAM requires user prompts for each image and instance.
This interactive design renders SAM unsuitable for high-throughput experiments, where large quantities of objects and images need to be segmented \cite{archit2025segment, Tsiporenko2024GoingBU}.
However, it appears promising to build a use-case specialist out of SAM, thereby keeping its knowledge while reduce the training costs and the required amount of labelled training data for one specific task. 

This idea has been investigated extensively~\cite{Na2024SegmentAC, archit2025segment, israel2023foundationmodelcellsegmentation,VandeLoo2025SAMCellGL}, 
resulting in different techniques for adapting SAM
for bridging the domain gap between the pre-training and target domain or to remove the need for user-defined prompts for each image.
Inspired by recent developments and their remaining short-comings, we propose prompt-tuning SAM, or short PTSAM, which leverages prompt-tuning techniques \cite{lester2021power, li2021prefix, Jia2022VisualPT} to fine-tune SAM's image encoder and mask decoder in an extremely parameter-efficient manner, thereby enabling robust training in data-limited scenarios and memory-efficient model distributions.
Our experiments on microscopic and medical imaging datasets underpin the following contributions of our approach:
\begin{enumerate}
    \item PTSAM allows for obtaining a specialist model with only \num{2048} trainable parameters that is on par with state-of-the-art microscopic image segmentation models.
    \item We find that additionally adapting the image encoder with PTSAM is beneficial for bridging the domain gap and further improves performance by up to 21\%.
    \item PTSAM provides consistent performance even when reducing the number of training images from \num{64} to \num{16}, leading to useful segmentation results where other methods that adapt the image encoder are prone to overfitting in a limited-data regime.
\end{enumerate}
These findings are visualized in Figure \ref{fig:dummy_res}.

\section{Related Work}

\subsection{Image Segmentation}
Without any doubt, the U-Net CNN architecture by \citet{Ronneberger2015UNetCN}
is the de-facto  standard 
for biomedical image segmentation.
A pivotal development of U-Net is nnU-Net \cite{Isensee2018nnUNetSF},
which provides automated configurations for hyperparameters of model and training, thereby enabling the development of solutions for segmenting medical images without advanced machine learning expertise.

CellSeg \cite{Lee2022CellSegAR}, on the other hand, pretrains a Mask R-CNN \cite{he2017mask} on nucleus segmentation tasks, to get an easy to adapt cell segmentation model.
Other approaches \cite{zheng2021rethinking,hatamizadeh2021swin,hatamizadeh2022unetr} extend convolutional neural networks with transformer layers \cite{Vaswani2017AttentionIA} to improve the performance for biomedical image segmentation.
However, these models need to be trained for each task individually.
\subsection{Promptable Vision Foundation Models}
In contrast to task-specific specialist models, 
foundation models \cite{Bommasani2021OnTO} provide the capability of being adaptable to tasks outside of their training domain.
One way for adaptation is prompting \cite{Brown2020LanguageMA, Achiam2023GPT4TR}, 
\eg, as well-known from language models like the GPT series \cite{Brown2020LanguageMA,Achiam2023GPT4TR}.
Following this success, transformers and self-supervised large-scale training were adapted for the vision domain \cite{Dosovitskiy2020AnII, radford2021learning, he2021maskedautoencodersscalablevision}.
A milestone of this research direction is the Segment Anything Model (SAM) \cite{Kirillov2023SegmentA},
a foundation model for image segmentation.
Based on user prompts such as point coordinates or boxes,
SAM adapts from segment-anything generalist to highly accurate instance-specific specialist.
However, needed promptability hinders applications in fully-automated workflows.

\subsection{Parameter-Efficient Fine-Tuning (PEFT)}
Historically, deep learning specialist models were either trained from scratch~\cite{Isensee2018nnUNetSF} or adapted through fine-tuning pre-trained parameters~\cite{he2016deep}.
However, 
for adapting a foundation model via training towards a specific downstream task, 
full fine-tuning is often not feasible,
especially because of the high computational requirements for training these models.
Therefore,
parameter-efficient fine-tuning (PEFT) methods have been developed~\cite{xu2023parameter, Han2024ParameterEfficientFF, Xin2024ParameterEfficientFF}, which update only a fraction of the trainable parameters or a small re-parametrization of the model.
A common PEFT approach is to select only a certain set of parameters to train.
Early approaches trained the last (few) layers of a network \cite{donahue2014decaf, long2015fully, howard2018universal},
where recently,
the set of these learnable parameters shifted towards bias terms \cite{cai2020tinytl}, 
normalization layers \cite{basu2024strong}, 
or weights in attention operations \cite{touvron2022three}.

In contrast to these selective fine-tuning methods,
feature adapters \cite{houlsby2019parameter, lin2020exploring, ruckle2020adapterdrop} introduce new parameters in the model
which are trained during training while the original network is kept frozen.
Similarly popular is low-rank adaptation (LoRA) by \citet{Hu2021LoRALA}.
LoRA freezes MLP layers and approximates their updates with a learnable low-rank matrix decomposition.
The lower the rank of these matrices, the lower the number of trainable parameters.
QLoRA \cite{dettmers2023qlora} further optimizes LoRA by quantizing the original weights to 4 bits.
A final notable technique is input tuning,
also known as visual prompt tuning,
\eg, by \citet{Jia2022VisualPT}, which adds new trainable parameters not directly in the model, but at the image space.
All these methods allow to adapt an existing model to a new task while keeping the overall input-output relation unchanged.
Thus, while applying PEFT to SAM would enable adapting SAM to a new domain~\cite{zhang2023customizedsegmentmodelmedical,archit2025segment},
they do not overcome SAM's needed promptability.

\subsection{Adapting SAM}
Different PEFT methods were developed for adapting SAM to a new domain.
In combination with removing the need of manual prompts, SAM becomes capable of automatically generating masks for certain tasks.
One approach to remove SAM's interactive design is to simulate user prompts.
\citet{wu2023selfpromptinglargevisionmodels} trained a logistic regression to predict foreground pixels on the image embeddings.
From this binary map, point and bounding box prompts are calculated, which are fed to the prompt encoder.
Methods for the microscopy domain \cite{archit2025segment, israel2023foundationmodelcellsegmentation, Na2024SegmentAC} use more complex autoprompting modules such as UNETR, DETR, and U-Net \cite{hatamizadeh2021unetrtransformers3dmedical, wang2022anchor, Ronneberger2015UNetCN}.
Additionally, these methods train further components of SAM to bridge the domain gap.
However, they still work with SAM's user prompts, which are obsolete when performing automatic segmentation.
Furthermore, to perform semantic segmentation, each instance needs to be segmented individually, which is inefficient for a large number of instances. 

In contrast, other SAM adaptations fine-tune the mask decoder to solve a specific task.
\citet{hu2023efficientlyadaptlargesegmentation} train different kinds of segmentation heads on top of SAM's image embeddings.
Their approach called AutoSAM uses SAM's mask decoder architecture and fully trains it for a specific task.
SAMed \cite{zhang2023customizedsegmentmodelmedical} additionally applies LoRA fine-tuning to the image encoder.
In the microscopic domain, SAMCell~\cite{VandeLoo2025SAMCellGL} fully fine-tunes the mask decoder as well, but also fully trains the image encoder.
While the number of parameters inside the mask decoder is manageable, it is still high enough to benefit from the use of PEFT techniques, which these works did not explore.

CellSeg1 \cite{zhou2024cellseg1robustcellsegmentation} heavily reduces the number of trainable parameters by using LoRA to parameter-efficiently adapt the mask decoder and the image encoder instead of fine-tuning them fully.
Hence, CellSeg1 accurately segments cells after training on just one input image.
However, to do that, users still need to provide prompts, \ie, it cannot segment objects fully automatically.

Our approach combines the advantages of PEFT and automatic segmentation, solving the issues of prompt generation and domain shift at once.
Instead of using LoRA such as most prior works, we adopt prompt tuning by \citet{Jia2022VisualPT} for adapting the mask decoder and image encoder, which is even more parameter-efficient.

\section{Method}
\begin{figure}
    \centering
    \includegraphics[width=\linewidth]{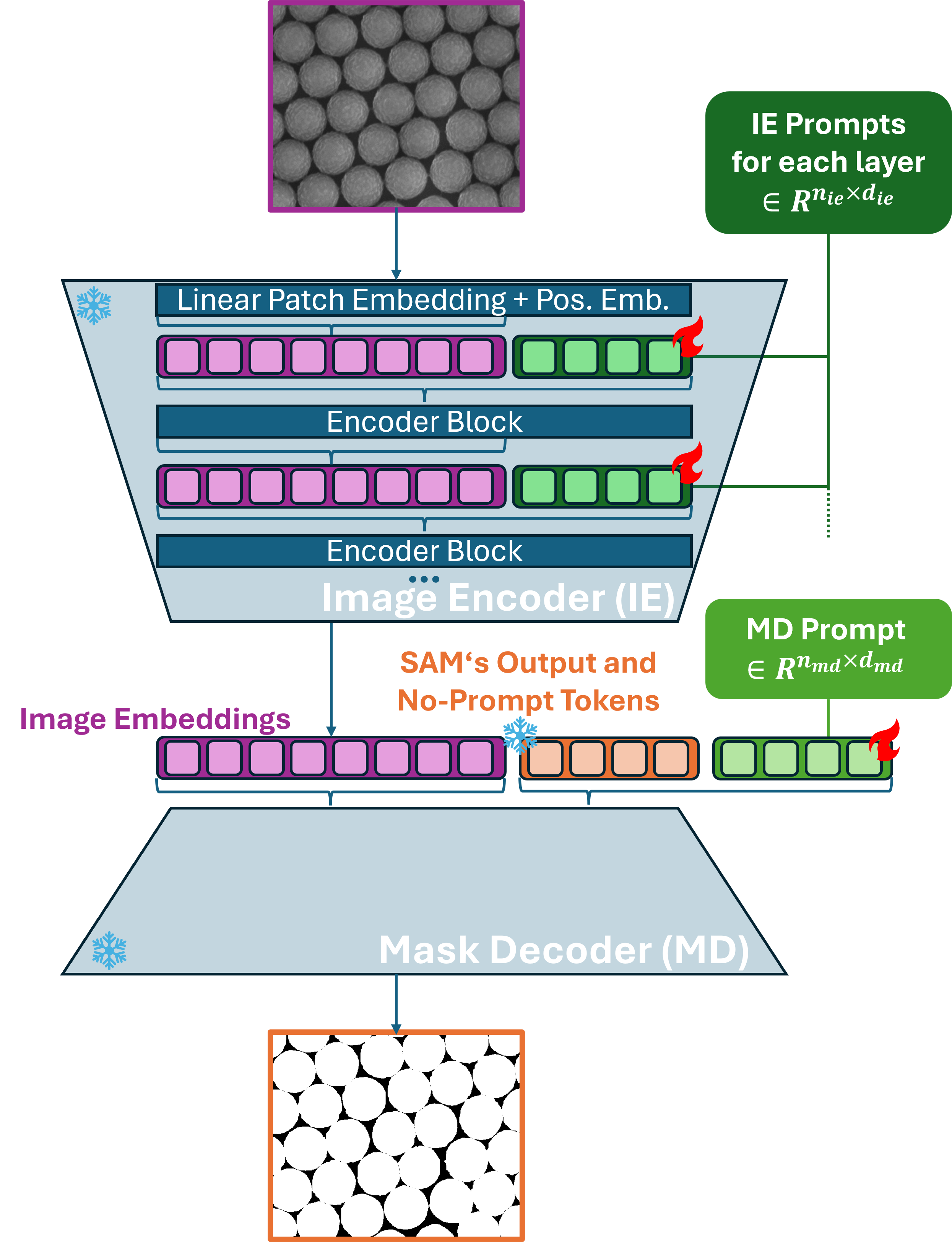}
    \caption{Overview of our PTSAM approach. We remove the prompt-encoder from the SAM architecture~\cite{Kirillov2023SegmentA}, which we keep frozen. Instead, we add learnable prompt parameters (green) into the mask decoder for becoming a use-case specialist and into the each layer of the image encoder to address domain shifts.}
    \label{fig:overview_ptsam}
\end{figure}

In the following, we first briefly describe the relevant parts of SAM's original architecture in \cref{subsec:sam} and then introduce our PTSAM approach for turning SAM into a specialist with few annotated examples in \cref{subsec:PTSAM}.

\subsection{SAM}
\label{subsec:sam}
As nicely visualized in \figurename~4 of \cite{Kirillov2023SegmentA},
SAM consists of three main components: the image encoder, the prompt encoder, and the mask decoder.
The image encoder of SAM is a Vision Transformer \cite{Dosovitskiy2020AnII}, pre-trained with a masked autoencoder (MAE) criterion \cite{he2021maskedautoencodersscalablevision}.
It takes input images with a size of \qtyproduct{1024 x 1024}{px}
and predicts an image feature map of resolution \qtyproduct{64 x 64}{px} with \num{256} feature channels per cell. 
To process high resolution input images,
the design for the image encoder is adopted from the work by 
\citet{li2022exploring} for vision transformers as object detection backbones.
In particular,
it uses a \numproduct{14 x 14} windowed attention and four global attention blocks, which are spaced equally throughout the layers of the transformer.

The prompt encoder embeds different types of user prompts, which are used to describe the desired object. 
Masks as dense prompts are added to the image embeddings through convolutional operations, while sparse prompts like points and boxes are fed to the mask decoder.
These sparse prompts are encoded with a combination of positional and learned embeddings. The positional embeddings encode the coordinates of the point or the corners of the bounding box.
The learned embeddings provide information about the type of prompt, \ie,
differentiating between the corners of the bounding box or whether the point belongs to foreground or background.
If no user prompts are provided, a no-prompt-embedding is used instead. 

SAM's mask decoder predicts the segmentation map based on two input streams: the image embeddings and a token stream.
The latter one consists of embeddings of the sparse user prompts and learned output tokens.
Both streams are the input to a two-layer transformer, which applies self attention to the token stream and cross attention from the token to the image steam and vice versa.
Finally, the image embeddings are upscaled with a transposed convolutional neural network and the output tokens are projected through a multi-layer perceptron.
The segmentation map is then calculated as a pixel-wise dot product of both results.
To account for prompt ambiguity, three output tokens are used to predict three output masks, one for object, part, and subpart.
A fourth IoU token is used for predicting which of these three outputs is the desired one.

\subsection{PTSAM}
\label{subsec:PTSAM}

Adapting SAM from a promptable generalist to a use-case specialist requires two things: 
i) an automatic generation of segmentation maps without user prompts and 
ii) an adaptation to a novel domain.
We propose to address both requirements with the same underlying base technique:
by transferring the idea of visual prompt-tuning \cite{Jia2022VisualPT} to SAM, giving rise to the name PTSAM.
Our resulting specialist adaptation of SAM to PTSAM  is visualized in \cref{fig:overview_ptsam}.

PTSAM keeps the original SAM architecture frozen and only trains inserted prompts.
Specifically, 
we add trainable prompt parameters into SAM's mask decoder,
thereby replacing the interactive segment \textit{anything} aspect of SAM with a trainable segment \textit{this specific task} ability.
Hence, SAM's prompt encoder becomes obsolete and, therefore, we remove it from the architecture.

In turn, we extend the token-stream used as input for the mask decoder with $n_\textrm{md}$ trainable prompts $\mathbf{p}_\textrm{md}\in\mathrm{R}^{n_\textrm{md} \times d_\textrm{md}}$, where the dimension of the mask decoder is $d_\textrm{md}=\num{256}$.
However, to maintain the pretrained behavior of SAM, we keep the no-prompt embedding in addition to all output tokens.
We only use the first of the three resulting segmentation maps as output.
During training, the added prompts are learned to segment the object of interest.
In contrast to the user prompts, which only encode positional information, our prompts belong to the latent prompt embedding space and
can hence be more expressive.

For further addressing any potential mismatch between the pretraining and target image domain, we additionally insert prompts into the image encoder using the deep visual prompt-tuning (VPT) method introduced by \citet{Jia2022VisualPT}.
Hence, instead of extending the inputs of the first layer only,
we prompt each layer $l$ of the vision transformer with $n_{ie}$ learnable prompts $\mathbf{p}_\textrm{ie}^{(l)}\in\mathrm{R}^{n_\textrm{ie} \times d_\textrm{ie}}$, with $d_\textrm{ie}$ being the dimension of the ViT.
Due to the windowed attention of the image encoder,
we insert the same prompts to each of the \numproduct{14 x 14} windows at a certain layer.
During training, we keep all of SAM's weights frozen, and learn only the inserted continuous prompt vectors $\mathbf{p}_\textrm{md}$ and $\mathbf{p}_\textrm{ie}^{(l)}$.
Thereby, user-prompt-free segmentation and domain shift can be tackled,
while needing only a small number of parameters to train.

\section{Experiments}
\label{sec:experiments}
We evaluate the effectiveness of PTSAM for transforming SAM into a use-case specialist on multiple application-relevant datasets,
specifically on three microscopic and one medical image segmentation task.
The performance is compared with state-of-the-art adaption techniques for SAM, as well as non-promptable segmentation standards. 
The choice of datasets allows for assessing the necessity of addressing the domain shift from natural to non-natural images. 

\subsection{Experimental Setup}
\label{subsec:exp-setup}

\subsubsection{Methods}
\label{subsubsec:exp-setup-methods}
We compare our PTSAM approach with two other methods for adapting SAM to specific downstream tasks.
Specifically known in the microscopy image analysis domain is 
CellSeg1 by  \citet{zhou2024cellseg1robustcellsegmentation}, which tunes the image encoder and mask decoder with LoRA \cite{Hu2021LoRALA}.
This method preserves the promptability of the model, making it not immediately applicable as use-case specialist.
Therefore, 
we modify CellSeg1 and 
use SAM's no-prompt-embeddings during training, enabling CellSeg1 to automatically generate masks.
Similarly popular is SAMCell \cite{VandeLoo2025SAMCellGL},
which however fine-tunes the entire image encoder.
When replacing full fine-tuning with LoRA training, 
SAMCell becomes a parameter-efficient adaptation technique, which is identical to a method known as SAMed \cite{zhang2023customizedsegmentmodelmedical} in the  medical domain.
We compare PTSAM against SAMed 
for a fair comparison.

We are specifically interested in analyzing the effects of training only the mask decoder (for making SAM a specialist) or additionally the image encoder (for addressing the domain shift).
Therefore, 
we analyze 
all SAM adapations in two settings: 
once with the parameter efficient fine-tuning of the image encoder, and once where we keep it frozen.
SAMCell's architecture with a frozen image encoder is the same as AutoSAM \cite{hu2023efficientlyadaptlargesegmentation}.

In addition to these SAM adaptations, we use nnU-Net \cite{Isensee2018nnUNetSF} as a baseline for training a specialist from scratch.

We give a summarizing overview on all these
methods in \cref{tab:meth_overview}, further highlighting which
parts are adjustable and the resulting number of trainable parameters. 

\begin{table}
    \centering
    \begin{tabular}{l|ccr}
         Method & IE & MD & \# Param \\
         \hline
         \multirow[c]{2}{*}{PTSAM (Ours)} & - & PT & \num{2048} \\
          & PT & PT & \num{75776} \\
         \hline
         \multirow[c]{2}{*}{CellSeg1 \cite{zhou2024cellseg1robustcellsegmentation}} & - & LoRA & \num{23552} \\
         & LoRA & LoRA & \num{171008} \\
         AutoSAM \cite{hu2023efficientlyadaptlargesegmentation} & - & Fully & \num{3645344} \\
         SAMed \cite{zhang2023customizedsegmentmodelmedical} & LoRA & Fully & \num{4212016} \\
         \hline
         nnU-Net \cite{Isensee2018nnUNetSF} & - & - & \num{59177872}\\
    \end{tabular}
    \caption{Methods overview with indications about the technique to train the image encoder (IE) and mask decoder (MD), and the corresponding number of trainable parameters.}
    \label{tab:meth_overview}
\end{table}

\subsubsection{Datasets}
We evaluate all
methods on three datasets from the lifescience microscopy domain as well as one medical image segmentation dataset.
Lifescience datasets where selected to cover different imaging modalities as well as different target applications. 
The \OrganoidBasic dataset \cite{van_beuningen_2023_10278229, Lefferts2024OrgaSegmentDB} was captured with bright-field microscopy and 
contains organoids to be segmented. 
As complementary modality, the \EMParticles dataset \cite{doi:10.1021/acs.jcim.0c01455} contains electron-microscopical images with varying targets to be segmented.
In addition, the \LiveCell dataset \cite{Edlund2021LIVECellALD} shows living cells in phase contrast images.
Finally, \KvasirSEG \cite{Jha2019KvasirSEGAS} serves as a non-microscopy dataset, covering  gastrointestinal polyp images as a representative scenario for medical imaging applications.
We include it in our experiments to analyze which approaches generalize beyond microscopy scenarios to other imaging and application domains. 

Since annotated
samples are expensive especially in these applications, the number of required images to train a specialist segmentation model should be as low as possible.
Hence, we evaluate each method in three scenarios with random subsets of \num{16}, \num{32}, or \num{64} annotated training samples. Across these scenarios, we keep consistent test sets with 20\% of all samples to allow for meaningful comparisons.

\subsubsection{Hyperparameters}
The scarcity of annotated training data poses another problem for real-world use cases: 
thorough hyperparameter optimization becomes infeasible.
Hence, we base our hyperparameter selection on common practices from the literature.
By keeping the setup consistent throughout different runs, we furthermore increase comparability.

For all SAM-based methods,
we use the ViT-b variant as the image encoder with the original SAM weights.
These weights are also used on the medical dataset instead of domain-specialized weights such as those from MedSAM \cite{ma2024segmentmedicalimages} because \citet{li2024adapting} found better performance when fine-tuning SAM's original weights to a new task.
We optimize the learning rate individually for each method on a heldout dataset.
For our PTSAM method, we found a learning rate of \num{0.05} to be optimal when keeping the image encoder frozen. Training the image encoder in addition reaches best metrics with a learning rate of \num{0.01}.

Method-specific hyper-parameters of other approaches, \eg, the rank in LoRA, 
are set to the values proposed in the original papers.
For our PTSAM model, we found in preliminary experiments that increasing the number of image encoder and mask decoder prompts beyond $8$ does not lead to further improvements (see \cref{subsec:abl}).
Hence, we set $n_\textrm{md} = n_\textrm{ie} = 8$.
With the mask decoder dimension $d_\textrm{md}=\num{256}$, we have $\num{256} \cdot 8 = \num{2048}$ trainable parameters for the mask decoder.
The ViT-b image encoder consists of $12$ layers, each having a dimension of $d_\textrm{ie}=\num{768}$, resulting in $12 \cdot \num{768} \cdot 8 = \num{73728}$ trainable parameters.

The well-tested setup used by \nnunet serves as a reference for other training parameters.
Specifically, we fix the number of epochs to \num{1000}
and the number of optimization steps in each epoch to \num{20}.
Thereby,
the number of training steps is independent of the dataset size, 
which allows conclusions about the relation of  dataset size and method performance.
Throughout the training, 
we reduce the initial learning rate with cosine annealing \cite{loshchilov2017sgdrstochasticgradientdescent}.
To prevent overfitting in our low-data regime, we apply heavy data augmentations.
We combine the setups from \cite{zhou2024cellseg1robustcellsegmentation, hu2023efficientlyadaptlargesegmentation, zhang2023customizedsegmentmodelmedical, Isensee2018nnUNetSF}, \ie, we apply randomized usage of rotation, cropping, elastic transformations, flips, Gaussian noise, and changes of Gamma, brightness, and contrast.
For the choice of the optimizer and loss, we follow other SAM adaptations \cite{israel2023foundationmodelcellsegmentation, zhou2024cellseg1robustcellsegmentation, VandeLoo2025SAMCellGL, hu2023efficientlyadaptlargesegmentation, zhang2023customizedsegmentmodelmedical} and use AdamW \cite{Loshchilov2017DecoupledWD} as the optimizer and a weighted sum of cross-entropy loss and Dice loss.
Following \citet{zhang2023customizedsegmentmodelmedical}, we weight these two losses with \num{0.2} and \num{0.8}, respectively.
Finally, we set the batch size to two,
which enables training SAMs image encoder even on consumer-grade hardware.

\subsection{Results}
\label{subsec:exp-results}

\begin{table*}
    \centering
    \begin{tabular}{lc|ccc|ccc|c}
        \toprule
         & & \multicolumn{3}{c|}{Frozen IE} & \multicolumn{3}{c|}{Trained IE} &  \\
         Dataset & \# Imgs & \textbf{PTSAM} & CellSeg1 & AutoSAM & \textbf{PTSAM} & CellSeg1 & SAMed & nnU-Net \\
        \midrule
        \multirow[c]{3}{*}{EMParticles} & 16 & $87.7_{\pm1.4}$ & $88.1_{\pm2.8}$ & \boldmath $\color{darkgreen}91.8_{\pm3.0}$ & $91.7_{\pm3.7}$ & $72.6_{\pm1.4}$ & $74.2_{\pm3.6}$ & $74.3_{\pm2.2}$ \\
        \cline{2-9}
         & 32 & $90.0_{\pm2.2}$ & $90.1_{\pm3.1}$ & \boldmath $\color{darkgreen}92.5_{\pm3.0}$ & $92.3_{\pm2.5}$ & $84.1_{\pm1.1}$ & $81.2_{\pm3.2}$ & $84.0_{\pm2.9}$ \\
        \cline{2-9}
         & 64 & $90.2_{\pm1.7}$ & $89.9_{\pm3.1}$ & $93.0_{\pm2.4}$ & \boldmath $\color{darkgreen}93.6_{\pm2.7}$ & $87.6_{\pm3.5}$ & $84.1_{\pm2.1}$ & $87.9_{\pm3.8}$ \\
        \cline{1-9} \cline{2-9}
        \multirow[c]{3}{*}{LiveCell} & 16 & $86.0_{\pm0.5}$ & $86.5_{\pm0.2}$ & $88.0_{\pm0.2}$ & \boldmath $\color{darkgreen}89.6_{\pm0.4}$ & $89.2_{\pm0.3}$ & $88.1_{\pm0.3}$ & $88.5_{\pm1.4}$ \\
        \cline{2-9}
         & 32 & $86.4_{\pm0.4}$ & $86.8_{\pm0.5}$ & $88.7_{\pm0.4}$ & \boldmath $\color{darkgreen}90.2_{\pm0.4}$ & $89.1_{\pm0.5}$ & $88.4_{\pm0.2}$ & $89.5_{\pm1.3}$ \\
        \cline{2-9}
         & 64 & $86.4_{\pm0.3}$ & $87.5_{\pm0.3}$ & $89.4_{\pm0.2}$ & \boldmath $\color{darkgreen}90.5_{\pm0.2}$ & $90.1_{\pm0.2}$ & $88.9_{\pm0.3}$ & \boldmath  $\color{darkgreen}90.5_{\pm0.3}$ \\
        \cline{1-9} \cline{2-9}
        \multirow[c]{3}{*}{OrganoidBasic} & 16 & $89.4_{\pm0.5}$ & $90.2_{\pm0.5}$ & $90.9_{\pm0.4}$ & \boldmath $\color{darkgreen}91.6_{\pm0.4}$ & $90.3_{\pm1.2}$ & $90.0_{\pm0.8}$ & \boldmath $\color{darkgreen}91.6_{\pm0.5}$ \\
        \cline{2-9}
         & 32 & $89.9_{\pm0.5}$ & $90.6_{\pm0.4}$ & $91.2_{\pm0.3}$ & \boldmath $\color{darkgreen}91.8_{\pm0.5}$ & $90.8_{\pm0.3}$ & $90.7_{\pm0.5}$ & $91.7_{\pm0.3}$ \\
        \cline{2-9}
         & 64 & $89.9_{\pm0.8}$ & $90.7_{\pm0.5}$ & $91.5_{\pm0.5}$ & \boldmath $\color{darkgreen}92.0_{\pm0.6}$ & $91.3_{\pm0.8}$ & $91.0_{\pm0.8}$ & $91.9_{\pm0.8}$ \\
        \cline{1-9} \cline{2-9}\\
        \cline{1-9} \cline{2-9}
        \multirow[c]{3}{*}{KvasirSEG} & 16 & $63.9_{\pm7.4}$ & $61.5_{\pm5.9}$ & $60.2_{\pm4.7}$ & \boldmath $\color{darkgreen}75.7_{\pm5.5}$ & $52.4_{\pm2.3}$ & $45.1_{\pm2.7}$ & $50.0_{\pm5.1}$ \\
        \cline{2-9}
         & 32 & $66.1_{\pm3.2}$ & $65.2_{\pm2.7}$ & $63.1_{\pm8.1}$ & \boldmath $\color{darkgreen}80.2_{\pm2.7}$ & $56.8_{\pm2.8}$ & $48.4_{\pm2.9}$ & $62.6_{\pm7.3}$ \\
        \cline{2-9}
         & 64 & $69.4_{\pm3.6}$ & $69.3_{\pm3.1}$ & $70.3_{\pm2.3}$ & \boldmath $\color{darkgreen}84.1_{\pm3.8}$ & $64.6_{\pm1.6}$ & $52.8_{\pm2.6}$ & $71.8_{\pm3.0}$ \\
        \bottomrule
    \end{tabular}
    \caption{Dice scores in percent of all experiments. We report mean and standard deviation obtained from three runs. Higher values are better. The method with  best mean score on each dataset is highlighted in \textbf{\textcolor{darkgreen}{green}}. The bold PTSAM method is ours.}
    \label{tab:results}
\end{table*}
We report all results with mean and standard deviation of Dice scores aggregated over three random splits per dataset in \cref{tab:results}.

\subsubsection{Lifescience Microscopy Image Segmentation}
On the microscopic datasets \EMParticles, \LiveCell, and \OrganoidBasic, our PTSAM with a trained image encoder reaches the highest Dice scores in  the clear majority of cases.
Noteworthily, training the image encoder with PTSAM consistently improves results on all datasets over prompt-tuning only the mask-decoder.
In contrast,
we observe on the \EMParticles dataset
that all other methods which adapt image encoders (including nnU-Net) fail to learn  reliabe segmentation models, most strongly visible with only \num{16} training images.
Given the large parameter counts, we attribute this behavior to overfitting.

When comparing the methods with a frozen image encoder, we initially observe that PTSAM leads to lower Dice scores than \cellseg and \autosam.
On a second view, however, 
we see that statistical significance of differences is often weak (as indicated by the relatively large standard deviations), and that in addition, the number of needed parameters for both methods is substantially larger.
Hence, we conclude that PTSAM allows training of microscopy specialists which are similarly accurate as other methods while needing only \num{2048} parameters to train.

\subsubsection{Medical Image Segmentation}
On \KvasirSEG, both PTSAM versions outperform their competitors in nearly all cases.
On closer look, 
we observe that all methods with a frozen image encoder (left column in \cref{tab:results}) perform comparably.
This finding is consistent with the previously discussed results on the microscopy datasets, and it is indeed surprising, given that our PTSAM uses  $>$\num{10}x less trainable parameters than the other methods.
When adapting the image encoder to account for the domain gap, however, PTSAM alone clearly benefits, while \cellseg, \samed, and even \nnunet obtain substantially worse results than with a frozen image encoder.
We again attribute the superior results of this larger PTSAM version to its robustness against overfitting.

\subsubsection{The Effect of Image Encoder Adaptation}
\begin{figure}
    \centering
    \includegraphics[height=5.5cm]{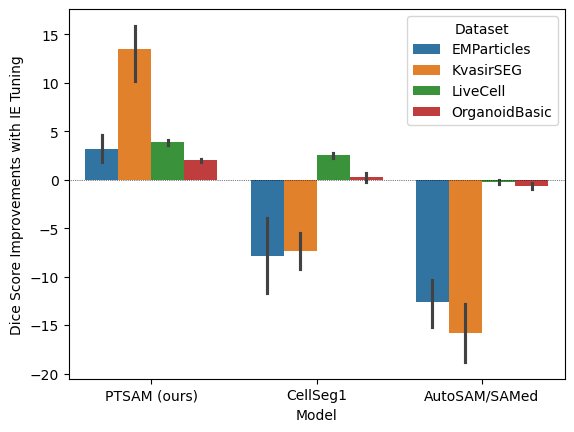}
    \caption{Comparing improvements of different methods when additionally tuning the image encoder. Higher values show that tuning the image encoder is beneficial.}
    \label{fig:ie_tuning_improvement}
\end{figure}

We have already seen on the medical dataset \KvasirSEG that 
other techniques which adapt the image encoder lead to inferior segmentation performance compared to their mask-decoder-only adapted versions.
Experiments on the \EMParticles dataset came to a similar conclusion.
We further analyze this observation by visualizing in \cref{fig:ie_tuning_improvement} the difference of the Dice score between models with and without an adapted image encoder.
Higher values indicate increased performance when training the image encoder.
In contrast to other SAM adaptations, PTSAM does not experience a performance decrease when tuning the image encoder.
Instead, we see consistent benefits of PTSAM with trained image encoder across all datasets. Interestingly, benefits are largest on medical data (shown in \textcolor{orange}{orange}). We assume that this improvement can be attributed to the application modality: objects in \KvasirSEG do not have clear edges, which is uncommon in SAM's original training domain \cite{hu2023efficientlyadaptlargesegmentation, ma2024segmentmedicalimages}.

\subsubsection{The Effect of Less Training Data}
\begin{figure}[t!]
    \centering
    \includegraphics[height=5.5cm]{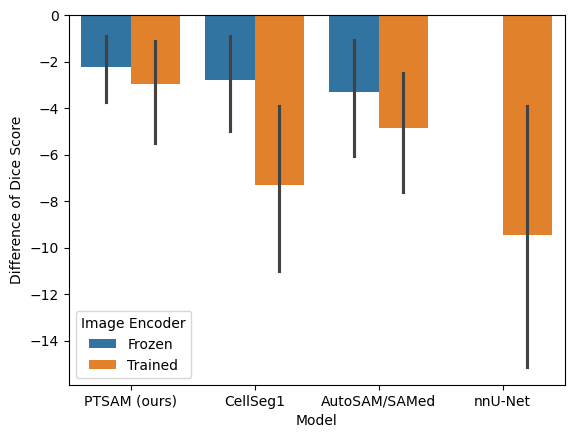}
    \caption{Performance decrease of the methods, when training with \num{16} instead of \num{64} images. Lower values indicate, that the method requires more training data.}
    \label{fig:diff_img_counts}
\end{figure}

In real-world scenarios,
even collecting just \num{64} images for a single use case can be intractable.
We therefore visualize the impact that a reduction of the training dataset from \num{64} to \num{16} samples has on the resulting segmentation accuracy of each method in \cref{fig:diff_img_counts}.
Results are obtained by averaging results from \cref{tab:results} of all four datasets.
We observe that when training without image encoder adaptation (\textcolor{blue}{blue}), segmentation accuracy drops only slightly by $\sim 2\%$ when reducing the training size by \num{4}x. Note again that \nnunet has no blue bar since the image encoder is always trained from scratch.
However,
when the image encoder is also adapted, 
we observe 
stronger drop in performance for the previous SAM adaptation methods of $\sim 4-6\%$ Dice score.
\nnunet struggles with a too small training set as well, leading even to an average drop of $\sim 8\%$ Dice score.
From all analyzed methods, 
our PTSAM is the least affected from small training set sizes, making it well-suited for few-shot scenarios.

\subsubsection{Qualitative Results}
\label{subsubsec:qualitative_results}
\begin{figure*}
    \centering
    \includegraphics[width=.93\linewidth]{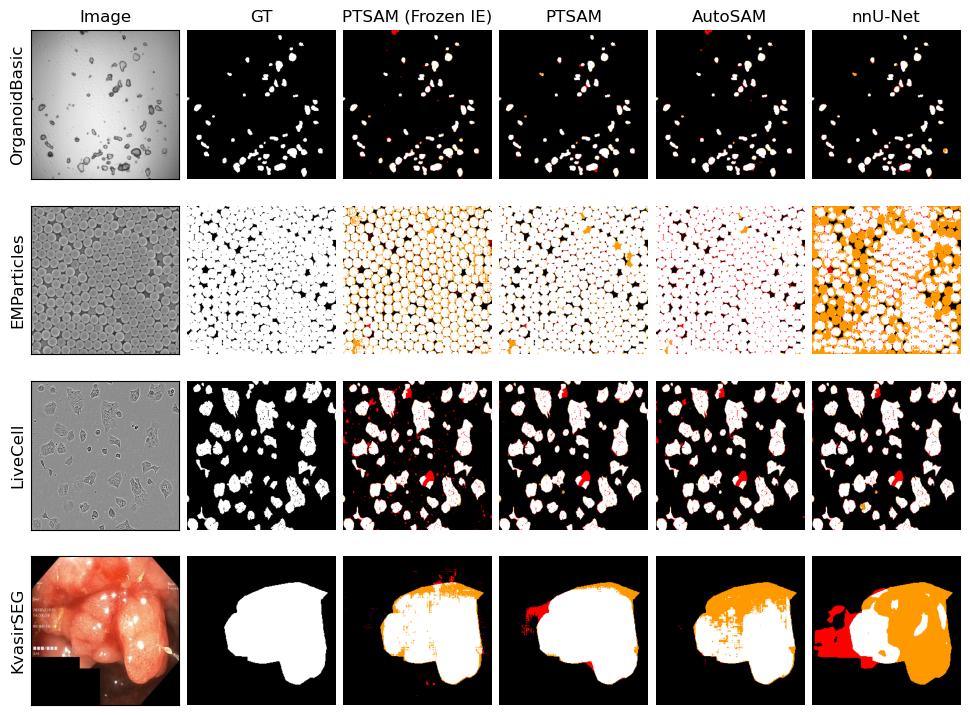}
    \caption{Qualitative segmentation results on three microscopy and one medical dataset (top to bottom). On each dataset, one example image along with its ground truth annotation as well as segmentation results of four chosen methods are shown from left to right. Results were obtained with models trained in a few-shot scenario with only \num{16} training samples. See \cref{subsubsec:qualitative_results} for details. \textcolor{red}{Red} areas indicate a false positive segmentation, whereas \textcolor{orange}{orange} indicates false negatives.}
    \label{fig:my_label}
\end{figure*}

We show random samples and the predictions of our PTSAM approach with frozen and tuned image encoder, \autosam as second best method, and \nnunet as classical baseline in \cref{fig:my_label}.
For better comparison, we highlight false negatives in \textcolor{orange}{orange} and false positives in \textcolor{red}{red}.
The segmentation maps on the \OrganoidBasic dataset are in all cases very promising and similar.
Noticeably, both methods with a frozen image encoder predict the same false positive object at the upper border of the image, while the fully trained PTSAM and \nnunet cannot find an organoid in the upper-left part of the image.
On the \EMParticles dataset, the resulting segmentation maps differ.
PTSAM with a frozen image encoder predicts most particles, while not being accurate on the boundaries.
With a tuned image encoder, the particles are segmented well but with some missing, which have a lower contrast in the input image.
\autosam predicts these but shows problems on the borders of the particles.
In contrast, \nnunet struggles a lot with the large number of particles.
On the sample from the LiveCell dataset, all methods behave similarly.
They predict the same false-positive structures, and they miss borders of individual cells within cell groups.
PTSAM without image encoder tuning also predicts multiple wrong small structures throughout the whole slice.
On the medical \KvasirSEG dataset, both PTSAM methods yield the most accurate segmentations.
PTSAM with a fixed image encoder leads to a patchy shape, while tuning the encoder results in a more consistent segmentation.
\autosam only segments the lower part of the polyp and the prediction of \nnunet is very off again.

\begin{figure}[b!]
    \centering
    \includegraphics[width=.9\linewidth]{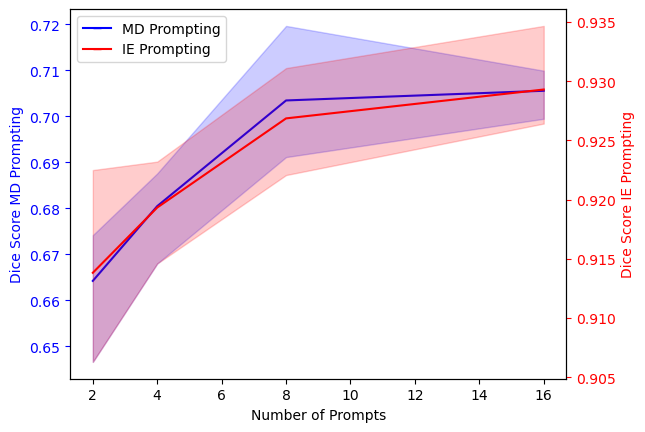}
    \caption{
    Ablation: investigating the effect of different numbers of prompts in mask decoder (MD) and image encoder (IE).
    For the \textcolor{blue}{blue} MD curve, the image encoder is frozen, meaning $n_\textrm{ie} = 0$. For the \textcolor{red}{red} IE curve, we use $n_\textrm{md} = 8$.
    Note the two y-axes scales.}
    \label{fig:abl}
\end{figure}

\subsection{Ablation on Number of Prompt Parameters}
\label{subsec:abl}
Our PTSAM method has two architecture-specific hyperparameters: the number of prompts for the mask decoder $n_\textrm{md}$ and the image encoder $n_\textrm{ie}$.
As described already,
all results in \cref{tab:results} were obtained by optimizing hyperparameters initially on a held-out dataset and then keeping them fixed for all remaining experiments.
The results of our initial ablation experiments are shown   in \cref{fig:abl}.
We observe that with a frozen image encoder, increasing the number of mask decoder prompts until $n_\textrm{md}=8$ leads to higher Dice scores, but reaches a plateau afterwards.
We then fix the number of mask decoder prompts to $8$ and additionally prompt-tune the image encoder with varying numbers of prompts.
Similar to the number of mask decoder prompts, inserting more than $8$ tokens only shows diminishing returns.
\section{Conclusion}
We presented prompt-tuning SAM (PTSAM),
a novel method for training a semantic segmentation model for data-limited non-natural image domains such as microscopy or medical image analysis.
PTSAM builds upon the strong segmentation capabilities of
Segment Anything (SAM) and leverages prompt-tuning
to adapt SAM to a new task and a new domain in a data-efficient and parameter-efficient manner.
Our approach solves two common problems of SAM for specialized applications: 
the necessity for spatial prompting for semantic mask generation,  
and the performance decrease under domain shifts.

We evaluated PTSAM on three microscopic and one medical dataset 
and compared it with three state-of-the-art methods for adapting SAM as well as \nnunet as the de-facto standard for semantic segmentation tasks.
We found that turning SAM from a generalist into a use-case specialist model only requires \num{2048} parameters, 
while providing segmentation accuracy
on-par with methods requiring $>\num{10}$x more parameters to adapt.
We observed that this parameter efficiency makes PTSAM robust to overfitting,
which allows for additionally adapting the image encoder with as few as \num{16} training samples.
While we found the segmentation accuracy of other methods decreasing when additionally adapting the image encoder, 
PTSAM's accuracy did not only not decrease but even increased further, even in such few-shot 
data-limited applications.
In summary,
we can conclude that PTSAM is an attractive combination of parameter efficiency and resulting segmentation accuracy for applications with limited training data and domain shifts.
We therefore see PTSAM an out-of-the-box solution for domain-specific image segmentation tasks, requiring only few annotated samples and no hyperparameters to tune.
{
    \small
    \bibliographystyle{ieeenat_fullname}
    \bibliography{main}
}

\end{document}